# Tailoring materials into kirigami robots

Saravana Prashanth Murali Babu, Aida Parvaresh, Ahmad Rafsanjani[*]

SDU Soft Robotics, Biorobotics section, The Maersk McKinney Moller Institute,
University of Southern Denmark, Odense 5230, Denmark

[*]Corresponding author: ahra@sdu.dk (www.softrobotics.dk), Date: June 25, 2024

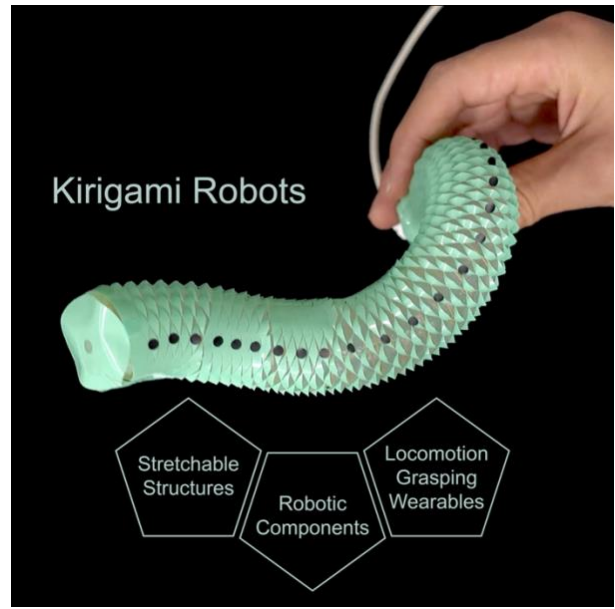

**THE BIGGER PICTURE:** As a design strategy, the kirigami technique can help bridge the gap between traditional rigid structures and soft, adaptable machines. By controlling the shape of cuts and folds, one can apply kirigami designs to any sheet of material across multiple length scales. In robotics, kirigami can help simplify robot designs by creating lightweight and compact actuators, sensors, power and control, and the skin and body of robots. Using kirigami manipulators in factory assembly lines can help handle delicate materials or construct intricate parts. Crawling kirigami robots can explore extraterrestrial planetary surfaces by tight packing during launch and deployment into a functional form after landing. Miniaturized kirigami robots can travel through the gastrointestinal tract to deliver drugs and perform biopsies on the spot. Moreover, imagine kirigami robots as integral components of smart cities, aiding disaster response to find people by swiftly adapting to diverse terrains and scenarios.

**SUMMARY:** Kirigami, the traditional paper-cutting craft, holds immense potential for revolutionizing robotics by providing multifunctional, lightweight, and adaptable solutions. Kirigami structures, characterized by their bending-dominated deformation, offer resilience to tensile forces and facilitate shape morphing under small actuation forces. Kirigami components such as actuators, sensors, batteries, controllers, and body structures can be tailored to specific robotic applications by optimizing cut patterns. Actuators based on kirigami principles exhibit complex motions programmable through various energy sources, while kirigami sensors bridge the gap between electrical conductivity and compliance. Kirigami-integrated batteries enable energy storage directly within robot structures, enhancing flexibility and compactness. Kirigami-controlled mechanisms mimic mechanical computations, enabling advanced functionalities such as shape morphing and memory functions. Applications of kirigami-enabled robots include grasping, locomotion, and wearables, showcasing their adaptability to diverse environments and tasks. Despite promising opportunities, challenges remain in the design of cut patterns for a given function and streamlining fabrication techniques.

## INTRODUCTION

Kirigami, a traditional paper-cutting craft, derives its name from the Japanese words '切る' (pronounced as "kiru," meaning to cut) and '紙' (pronounced as "kami", meaning paper).[1] This art form has gained popularity for scientific applications due to its versatility and ability to create intricate three-dimensional shapes. The incorporation of cuts imparts structural properties and complex behaviors to a solid sheet of material. If we tessellate cuts on a plain sheet, a kirigami metamaterial emerges in which collective opening and local deformation of ligaments transform it into a uniformly textured surface.[2-4] Encapsulating a soft inflatable actuator with a kirigami sheet affects its morphing behavior by preferentially directing internal forces during deformation[5]. Asymmetric popups in kirigami metasurfaces can generate anisotropic friction,[6] and introducing kirigami into adhesive films enhances film adhesion.[7,8] These emergent behaviors enable new robotic functionalities for locomotion, grasping, and wearables. While recent comprehensive reviews cover broader applications of kirigami structures in various domains,[9-12] this Perspective will focus on how kirigami can foster functions in robotic systems.

Kirigami distinguishes itself from paper-folding art of origami by incorporating cuts, and while fold may also exist, they are not essential. As a structural modification, kirigami does not inherently possess any robotic functionality. However, exploiting kirigami principles in designing robotic elements, such as actuators, sensors, batteries, and flexible electronics, can elevate the morphing and adaptability of robots. In soft robots where physical compliance is pivotal in shaping their embodied intelligence, kirigami can multiplex programmable deformations with various emerging functionalities.[13] From a sustainability standpoint, since the properties of kirigami structures purely rely on cut geometry[14], it allows for utilizing environmentally friendly alternatives in fabricating robots, replacing non-recyclable materials, and eventually reducing their environmental footprint.[15] Finally, the mechanical response of kirigami is essentially size-independent, making them applicable to electromechanical systems at micro (MEMS[16]) and nano (NEMS[17]) scale.

## KIRIGAMI ENHANCES STRETCHABILITY

From a mechanics perspective, the deformation of kirigami structures primarily relies on bending rather than stretching, which requires minimal stored elastic energy.[18] This provides a cost-effective method for shape morphing under small actuation forces. Consequently, the bending-dominated response of kirigami structures extends their stretchability beyond the fracture strain of the base material, making them more resilient to tensile forces.[19,20] Optimizing cuts helps distribute stresses more uniformly and reduce stress concentration by enhancing compliance, which reduces the risk of structural failure and increases the longevity of the robot.

The ability of kirigami structures to morph, stretch, and conform to diverse surfaces while preserving durability and energy efficiency renders them a compelling platform for robotic solutions. Thin, flat kirigami sheets made of plastics, polymer composites, textiles, or even metallic foils can morph into three-dimensional space through out-of-plane rotation of hinges.[21] When designing thin kirigami sheets, it is crucial to consider their inherent out-of-plane symmetry. An imperfection such as a supporting backing layer,[22] curvature,[23] or additional engraved slits[24] ensures predictable pop-ups in the desired direction. Despite myriad cut patterns (see Figure 1A-D), the characteristic force-displacement of most stretchable kirigami structures features three distinct regimes.[25] At small deformations, their response is linear and stiff; during the opening of the cuts at moderate to large elongations (>100%), they become nonlinear and soft, and when the cuts are fully open, stretch-dominated deformation prevails, the structure hardens again and eventually breaks with fracture failure.

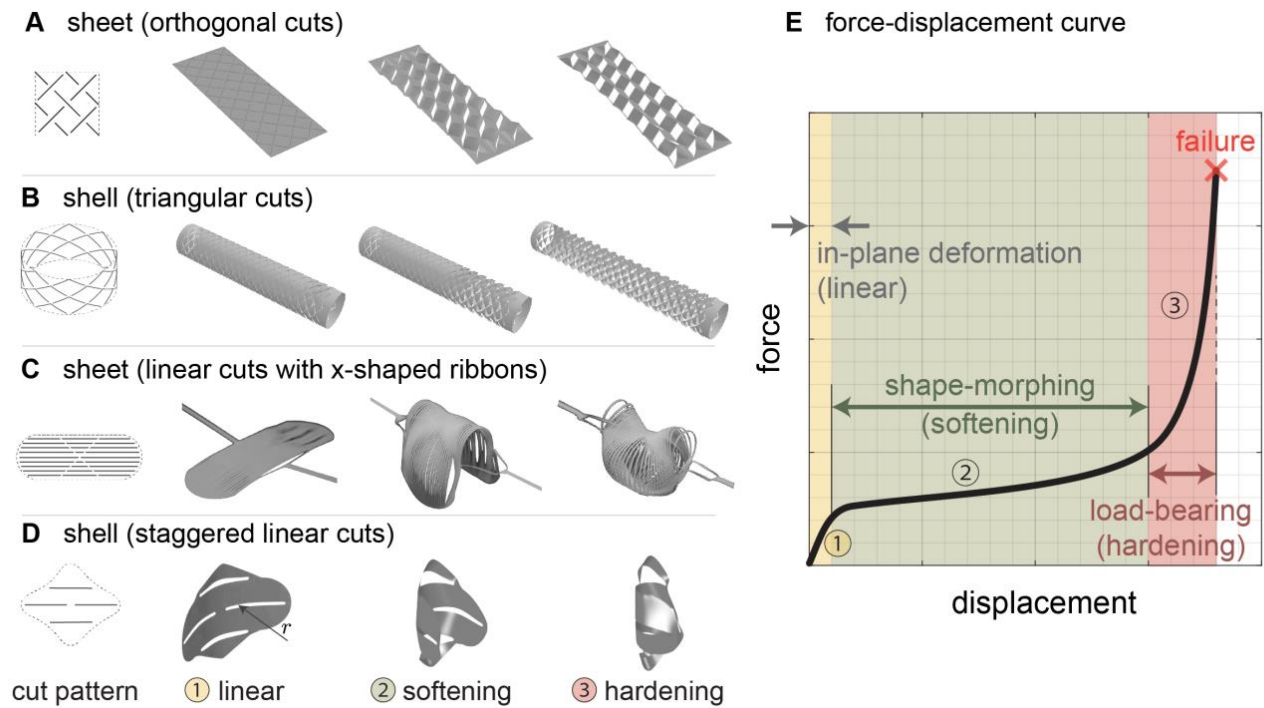

Figure 1. Mechanical response of stretchable kirigami structures.
(A) kirigami sheet with orthogonal cuts.[2]
(B) kirigami shell with triangular cuts.[23]
(C) kirigami sheet with linear cuts and X-shaped ribbons.[15]
(D) kirigami shell with staggered linear cuts.[50]
(E) The typical force-displacement curve of stretchable kirigami structures consists of three regimes: linear, softening, and hardening with indication of the morphing and load-bearing regions.

The initial linear regime of the force-displacement response often diminishes for thin kirigami sheets and shells due to the plasticity of cut corners upon the first loading cycle. Consequently, in subsequent cycles, they typically transition from the flat state into the popped-out, nonlinear, soft regime as depicted in Figure 1E. The *softening regime* is particularly advantageous for integration into robotic systems. With only a tiny tensile force, these sheets undergo deformation and shape transformation, making them suitable as flexible sensors or conformal end effectors.

The *hardening regime* is beneficial for applications requiring load-bearing capabilities. However, the base material must withstand forces to exploit this regime without reaching its fracture strain. Moreover, the J-shaped force-displacement curve of thin kirigami structures is akin to that of biological tissues such as skin, ligaments, and blood vessels in which a natural *strain-limiting* mechanism prevents damage from excessive strain.[26] This similarity suggests utility for kirigami in biomedical devices, allowing for the integration of stretchable electronics with host tissues or organs.

**ROBOTIC KIRIGAMI COMPONENTS**

Sensing, planning, and action are perhaps the three most important functionalities in robotics.[27] Kirigami serves as a versatile design framework applicable to robotics, facilitating the development of lightweight, adaptive, and easily fabricable robotic components.[28] For example, researchers have proposed leveraging stretchable kirigami components for layer-by-layer manufacturing of composite mechanisms, streamlining the fabrication of mesoscale customized robots.[29] They have demonstrated the effectiveness of this approach by creating a foldable composite inchworm robot with three kirigami layers, serving as an actuator, sensor, and contact pad with directional friction (Figure 2A). Such developments at the system level underscore the potential of kirigami for advancing robotics technology.

**Kirigami actuators**— An actuator is a device that converts input energy into mechanical energy to produce force, torque, or displacement. Incorporating cuts into the actuator design enables various deformation modes, including stretching/contracting, bending, twisting, rolling, and combined modes.[18] These movements can be triggered by utilizing various energy sources. For example, mechanical deformations directly applied to stretchable kirigami structures generate complex motions that can be preprogrammed through cut patterns. Alternatively, pneumatically-driven elastomeric inflatable structures covered with kirigami metamaterial skins enables the modulation of the robot's surface texture for precise control over frictional interactions.[30] Kirigami sleeves can define the deformation of pneumatic actuators (Figure 2B), allowing them to morph and conform to complex geometries.[5,31] Laminated kirigami structures with embedded air pouches (Figure 2C) can generate large contractions to mimic the movements of the human muscles.[32]

Electrical stimulus allows more dynamic, precise, and programmable control over kirigami shape transformations. For instance, kirigami actuators made of shape memory alloys (SMA) with simple line cuts can be directly driven through Joule heating. They cool rapidly due to the large surface area, and are compatible with laser micro-machining, making them ideal for manufacturing mesoscale composite robots.[29]

Integrating smart materials into kirigami structures allows for the utilization of external stimuli to wirelessly control the motion of untethered robots, e.g., with magnetic fields,[33,34] light,[35] and chemical potentials (e.g., relative humidity[36]). One example is magneto-responsive kirigami actuators made of a silicone matrix embedded with NdFeB particles that are magnetized to saturation under an impulse magnetic field ($\approx 3T$) to align their magnetic pole directions. Then, applying a small magnetic actuation field ($\approx 150 mT$) post-fabrication triggers buckling, stretching, and folding.[37] This approach enables versatile 2D-to-3D and 3D-to-3D shape changes under external magnetic fields (Figure 2D). Another stimuli-responsive material is Liquid Crystal Networks (LCN), which can respond to heat due to molecular alignment/order changes[38]. Light-responsive kirigami film actuators were created by infiltrating LCNs with light-sensitive elements such as photoswitches or photoabsorbers. The molecular orientation is conveniently aligned across the kirigami film thickness. Upon irradiation with specific light wavelengths, the film can be reshaped via out-of-plane bending into various deformation modes, including stretching, twisting, and rolling (Figure 2E). Another study demonstrated light-responsive kirigami pop-ups made of hygroscopic bilayers[36]. For the active layer, vanadium oxide nanowires were directly grown on a cellulose fiber network to increase the surface area by 30 folds and boost hydrophilicity for achieving fast, reversible actuation upon shining lights.

**Kirigami sensors**— A proprioceptive sensor enables a robot to gauge its own physical state relative to its surroundings, while an exteroceptive sensor detects external influences on the robot. Both sensor types are essential for informing robots for decision-making and adaptation to different environments. However, a challenge in developing soft sensors is the conflict between electrical conductivity and compliance because most conductive materials are not stretchable[39]. Stretchable kirigami structures can bridge this gap by turning an electrically conductive sheet into a stretchable structure while keeping mechanical stresses within a safe zone[40]. Therefore, we can benefit from a wide range of inextensible electrically conductive materials for fabricating sensors such as poly(vinyl alcohol)/graphene oxide nanocomposite[19] and laser-induced graphene[41].

Researchers developed a series of piezoresistive kirigami sensors (Figure 2F) by utilizing off-the-shelf sheets of electrically conductive silicone that were plasma bonded onto the surface of pneumatic elastomeric actuators[42]. They fed the acquired data into a deep learning algorithm based on the Long Short-Term Memory (LSTM) technique to enable distributed proprioception, facilitating the estimation of the complete 3D continuum shape of a sophisticated, multi-segment soft robot arm. Other researchers fabricated a stretchable kirigami sensor from composite layers of metallic foils and polymers[29]. They harnessed the increase in electrical resistance caused by the local tensions developed during the out-of-plane regime of the stretchable kirigami to measure the elongation at two ends. Notably, the initial in-plane deformation of the kirigami sensor is insensitive to small deformations, and the plastic deformation in the first cycle may result in drift in measurements. However, the response of these sensors is consistent over repeated subsequent cycles. Integrating two sensors into a mesoscale inchworm robot enabled controlled locomotion with angular proprioceptive feedback.

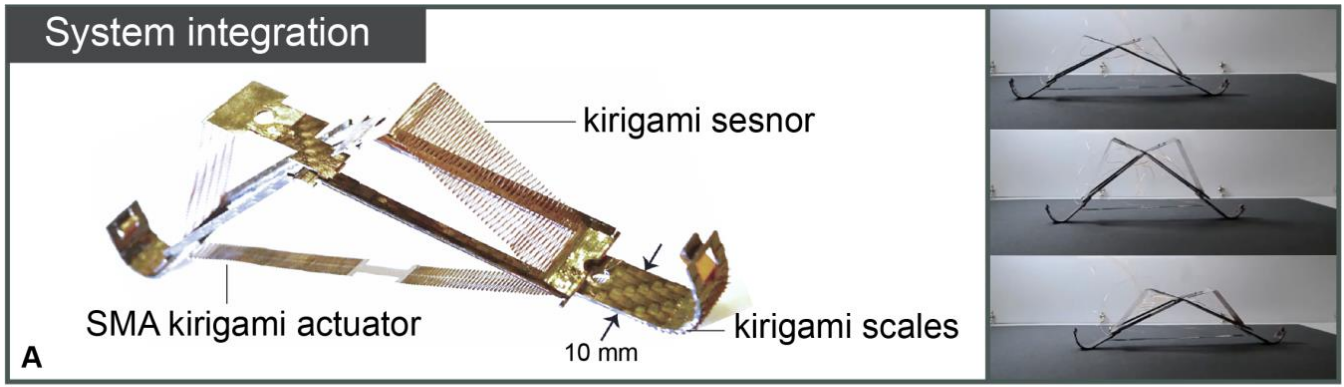
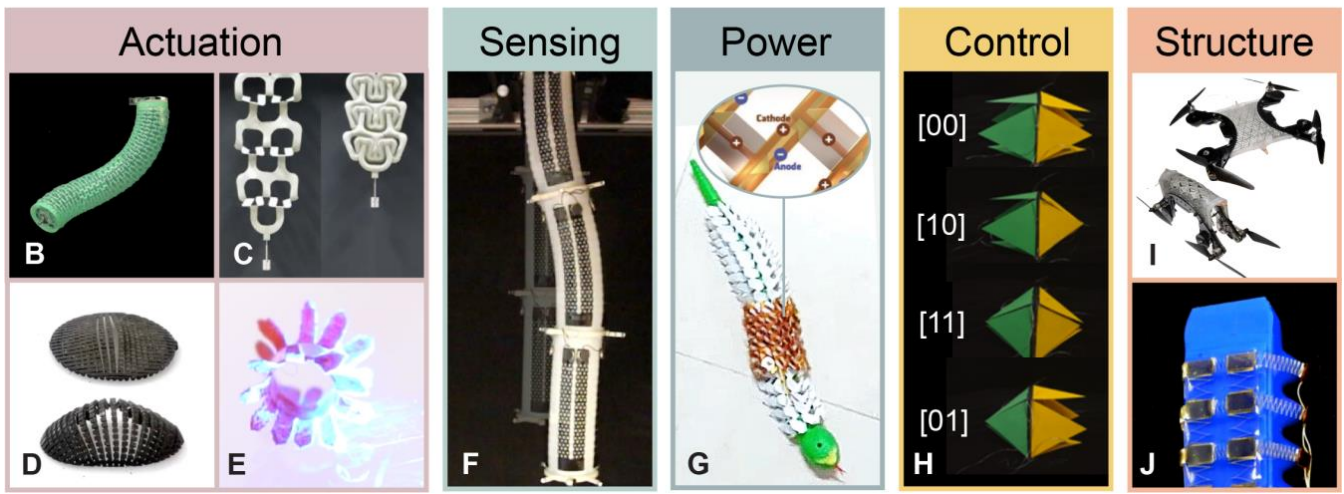

Figure 2. Kirigami-inspired components for robots.
(A) System integration.[29]
(B-E) Kirigami-enabled actuators, (B) inflatable with kirigami skin[5], (C) laminated kirigami motor pouches,[32] (D) magneto-responsive[37] and (E) light-responsive.[38]
(F) Proprioceptive kirigami sensor.[42]
(G) Kirigami scale battery.[45]
(H) Kirigami controller.[46]
(I) Shape morphing kirigami body for a drone-to-wheeled transformable robot.[47]
(J) Kirigami interconnections for a sensorized soft robotic finger.[48]

Reflex-driven pathways that enable living organisms to sense and respond to their environment inspired researchers to devise a somatosensory light-driven robot (SLiR) by incorporating kirigami into heat- and light-responsive thin-film made of ferroelectric poly (vinylidene di-fluoride) (PVDF), photothermal polydopamine reduced graphene oxide (PDG) and conductive graphite–carbon nanotube composites (graphite–CNT). The PDG absorbs and converts light into heat, and its contrasting coefficient of thermal expansion with PVDF makes the thin film a bimorph with light-induced actuation. The thin-film layup of active materials enables simultaneous yet non-interfering perception of the robot's actuation deformation state and body temperature utilizing respective piezoresistive and pyroelectric effects[43].

**Kirigami battery**— Most untethered robots rely on batteries for energy storage to power their operations. The utilization of shape-morphing kirigami structures allows energy sources to be integrated directly into the structures of robots – known as embodied energy[44] – without requiring separate battery packs. Inspired by the scales of a snake, researchers fabricated a shape-morphing battery (Figure 2G) using aluminum and metal foils cut into specific patterns and folded along predefined lines[45]. The battery integrates hexagonal cells comprising a cathode of lithium cobalt oxide-coated aluminum foil, a graphite-coated copper anode, a separator, and an electrolyte solution of $LiPF_6$. This battery underwent *in situ* electrochemical testing involving cyclic mechanical loading to assess its charge-discharge performance and electrochemical characteristics under dynamic conditions, such as stretching and bending. The battery's multi-axial deformability enables it to conform to various shapes and withstand complex meandering movements.

**Kirigami controller**—The opening and closing of kirigami cuts can mimic a mechanical bit, enabling the realization of mechanical computations. Researchers have developed a kirigami interactive mechanologic coupled with a self-powered triboelectric nanogenerator (TENG) mechanoreceptor through a signal transmission/processing module, which triggers the folding and unfolding of motorized hinges of a pyramid-shaped kirigami geometry (Figure 2H)[46]. This geometry possesses bistable resistive states thanks to integrated photoresistors, representing the mechanologic concept. This approach can also help realize mechanical registers, counters, and memory functions.

**Kirigami body**— Integrating kirigami into robot bodies can make them lightweight, flexible, and compact, with programmable kinematics and shape-morphing behavior. Researchers created a kirigami structure with reversible plasticity and polymorphic reconfigurability to rapidly change shape (< 0.1s) across negative, zero, and positive Gaussian curvature surfaces.[47] The multilayer structure encapsulates a rigidity-tuning kirigami endoskeleton of a low–melting point alloy (LMPA) within an elastomeric kirigami layer. Additionally, an embedded electrically driven liquid metal heating layer transforms the LMPA from a high elastic modulus solid to a flowable liquid through Joule heating at relatively low melting temperatures (62 °C), facilitating fixing load-bearing configurations and reversible shape relaxation. This kirigami composite enabled a flying drone to reconfigure into a driving vehicle on demand, overcoming limitations and trade-offs between extensibility and load-bearing capacity in morphing robots while eliminating power requirements to sustain reconfigured shapes (Figure 2I). Another study introduced a robotic skin designed for a soft robotic finger, featuring a flexible and stretchable multi-sensing array capable of temperature and proximity sensing.[48] The sensors are interconnected by kirigami wire traces embedded within a soft and stretchable silicone layer. The serpentine kirigami interconnections stretch as the fingers open and close, enabling expansion without significantly impacting sensor readings (Figure 2J). In another study, researchers harnessed the viscoelastic behavior to achieve propagating kinks in purely dissipative kirigami.[49] They showed that if a viscoelastic kirigami is stretched fast, after a while, the resulting pop-ups eventually snap from one texture to another configuration, enabling machine-like functionalities, such as sensing, dynamic shape morphing, transport, and manipulation of objects.

**ROBOTIC APPLICATIONS**

In this section, we highlight some areas of application that have benefited from the adoption of kirigami into robotics by showcasing state-of-the-art kirigami-enabled robots (see Figure 3).

**Kirigami Graspers—** Deforming a kirigami structure has two implications that can enhance contact with arbitrary objects and make grasping more effective. The introduction of cuts boosts the conformation of the gripper structure to complex shapes and enhances its surface roughness through local deformations. This combination has powered various shape-morphing kirigami grasping systems that enable the universal pick-and-place of objects with diverse shapes and stiffness.

Researchers proposed a gripper realized through laser-cutting a two-dimensional kirigami pattern of staggered lines onto a polyethylene terephthalate (PET) thin elastic sheet, which is subsequently heat-pressed onto a cylindrical substrate to attain its curved structure[50]. The kirigami shell is material-independent and scalable and can be connected in series and in parallel and be actuated by stretching the edges to handle objects of various shapes and sizes (Figure 3A). In a related design, a multilayer kirigami structure was created by encapsulating a perforated PET film within a silicone layer[51]. This structure, conversely, can be actuated by compressing side edges and can grasp everyday objects, including challenging ones such as a hammer and a wrench (Figure 3B). Studying the geometric mechanics of disordered kirigami has led researchers to control the deployment trajectory of a kirigami sheet consisting of a single slit perpendicular to the loading axis and a few lateral cuts[52]. This sheet serves as a robotic gripper capable of holding objects passively even in the absence of applied force when the cut boundary relaxes to contact the object (Figure 3C). Other researchers exploited a mathematical theory that correlates the geodesic curvature along the boundary with the Gaussian curvature to program the global shape of a stretchable thin kirigami sheet comprised of a series of long parallel cuts[53]. They showed that kirigami sheets with positive, zero, and negative boundary curvatures give rise to sphere, cylinder, and saddle shapes, respectively, upon stretching. Then, they created a multisegmented monolithic gripper by separating cuts with a central X-shaped uncut region that formed a saddle in the middle and convex surfaces at the extremities[15]. The morphological adaptability of this structure enhanced its performance in handling delicate and challenging objects such as egg yolk, foams, and human hair, as well as ultrathin, sharp, and heavy items (Figure 3D). They also demonstrated its material independence and potential for reducing the ecological footprint by crafting this gripper from a dracaena leaf (Figure 3E).

**Kirigami Locomotors**— When equipped with an actuation mechanism, kirigami can function as a reconfigurable skin or a flexible body to facilitate the locomotion of robots.

A snakeskin-inspired stretchable prismatic kirigami skin, wrapped around a simple extending soft actuator, enabled rectilinear locomotion under cyclic inflation and deflation[30]. Upon stretching, the asymmetric kirigami scales popped out in a preferred direction, creating a directional friction contrast that was considerably higher in the backward direction than in the forward direction. In each cycle, when the actuator elongated, the kirigami skin established an anchor point with the ground near the robot's tail, pushing the head forward. Conversely, when deflated, the anchorage shifted to the front, pulling the tail forward and completing one stride (Figure 3F). Subsequent work showed that the pop-ups in a rolled kirigami propagate from one end to another. Locally thinning the hinges at two extremes triggered pop-ups at two ends, and the resulting non-uniform deformation improved the crawler's grip on the substrate, minimized backslide, and increased crawling speed[23].

Integrating kirigami skin onto two bending actuators attached in series enabled the emulation of snake's lateral undulation.[54] Arranging kirigami cuts in a curvilinear lattice altered the initial orientation of the scales, allowing the snake robot to slither faster.[55] This design facilitated the lateral rotation of scales during bending, resulting in a reduction in the anisotropic friction ratio as fewer scales aligned along the spine (Figure 3G). This led to higher lateral friction, offering increased lateral resistance, which is crucial in generating undulating locomotion in snakes. Moreover, a kirigami skin with threefold symmetry, attached to a stretchable backing layer, adjusts according to the loading direction when manipulated by pairs of extending actuators positioned on the perimeter of a triangular-shaped robot[22]. This reconfiguration results in the creation of an asymmetric popup effect, enabling customizable friction patterns in three directions and facilitating in-plane steering (Figure 3H). In another study[56], researchers mimicked the two-anchor crawling locomotion of earthworms in loose soil environments by incorporating a radially expandable kirigami skin onto the first and last actuators of a three-segment soft robot (Figure 3I). Compared to a bare earthworm robot, the kirigami skinned robot exhibited a larger maximum drag force, and higher traction in cohesive soil terrain, and greater forward displacement per gait cycle. Moreover, adding folds to kirigami metamaterials based on rotating squares and triangles enhanced their reconfigurability by increasing the rotation range of units, enabling a soft robot to turn on the spot without requiring maneuvering.[57]

By incorporating magnetic particles into an elastomeric kirigami structure, researchers developed an untethered crawling robot for deploying a sensor attached to it in the human body (Figure 3J) as an alternative to conventional catheter- or probe-based sensing methods[58]. After laser cutting, they programmed an anisotropic magnetization profile using a Halbach array into the structure by pre-stretching it during the magnetization process. (A Halbach array refers to an arrangement of magnets that enhances the magnetic field on one side while nearly canceling it on the opposite side.) The robot was actuated remotely by a permanent magnet attached to a robotic arm that, when approached by the magnet and distanced from it, could stretch and release the kirigami and move it forward with directional friction induced by asymmetric pop-ups. By varying the magnetic field, the robot could steer and flip direction, cross gaps, move on dry and wet surfaces, climb vertically, and perform inverted crawling. Researchers utilized kirigami cuts to create a centipede robot based on SliR design[43] (also mentioned in the section Kirigami sensors). Light-induced reconfiguration of the body shape and frictional anisotropy of the feet lead to directional locomotion that can be further controlled by shining localized light pulses (Figure 3K).

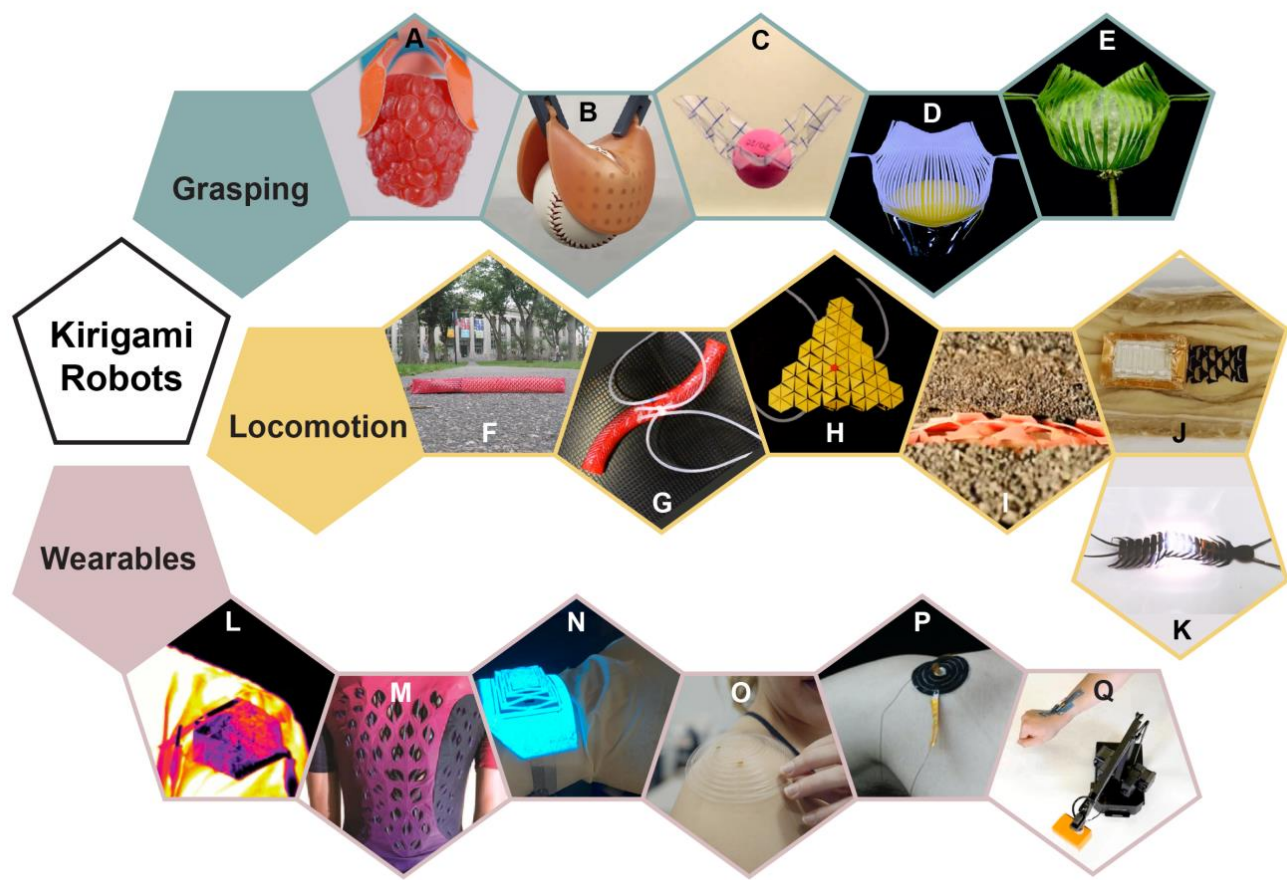

Figure 3. Applications of kirigami-enabled robots in grasping, locomotion, and wearables.
(A-E) Grasping with kirigami shells (A) under tension,[50] (B) under compression,[51] (C) passively,[52] (D) with programmed ribbons,[53] (E) using sustainable materials.[15]
(F-K) Mobile kirigami-enabled robots capable of (F) rectilinear locomotion,[30] (G) lateral undulation,[55] (H) planar steering,[22] (I) burrowing,[56] (J) magnetic actuation,[58] (K) light-induced motion[43].
(L-Q) Kirigami-enabled wearables for (L) thermoregulation,[59] (M) sweat control,[60] (N) electroluminescent durability,[61] (O) multiaxial motion capture,[62] (P) thermotherapy,[41] (Q) human-robot-interaction.[63]

Kirigami shows promise in enabling system-level integration for robotic locomotion (Figure 2A). Researchers have devised a layer-by-layer manufacturing process to seamlessly incorporate various kirigami components – including a stretchable sensor, a shape memory alloy (SMA) kirigami actuator, and passive kirigami scales – into a mesoscale crawling robot capable of inchworm-like motion[29]. This robot's locomotion is finely tuned, relying on the equilibrium between the moments generated by the SMA actuator and the elastic energy stored in stretchable sensors. It harnesses kirigami-enabled proprioceptive feedback and anisotropic friction, thus enabling precise and controlled movement.

**Kirigami Wearables**—Researchers developed an electrically driven, kirigami-enabled electrochromic thin-film device that can effectively tune the thermal radiation heat loss of the human body[59]. The kirigami design provides stretchability and conformal deformation under various modes, and electronic control enables programmable personalized thermoregulation (Figure 3L). The device consists of a working/counter electrode pair, and both are electrodeposited polyaniline (PANI) on an Au-sputtered nano-porous nylon membrane. In the working electrode, the gold layer acts as a strong reflector, and the electrochromism in the M-IR window allows for tuning of emissivity to stabilize the radiative heat loss. In another work, researchers engineered the hygroscopic and biofluorescent behaviors of living cells to design biohybrid wearables (Figure 3M), which give multifunctional responsiveness to human sweat[60]. In this design, a series of initially flat biohybrid flaps cut into a garment based on heat maps and sweat maps of the back that open and regulate ventilation and body temperature when exposed to skin with high humidity.

Using a thermally conductive cellulose nanofiber film, researchers leveraged kirigami to develop a flexible cooling system through convective heat dissipation (Figure 3N) that can increase the longevity of electroluminescent wearable devices and control the emission-angle range[61]. Strategically integrating strain gauges into a planar rotationally symmetric kirigami resulted in a wearable, conformal patch (Figure 3O) that can capture the multiaxial motion of shoulder joints and muscle behavior in real time[62]. A similar kirigami structure was utilized by other researchers to create stretchable laser-induced graphene heaters (Figure 3P) using a single-step customizable laser fabrication method for wearable thermotherapy.[41] Researchers programmed nonlinear open polygonal-shaped cuts into an elastomeric adhesive film to create strong yet releasable adhesives for wearable devices such as gloves[63]. When shearing the kirigami adhesive along the cut base, cracks propagate backward, resulting in higher adhesion and stronger grip whereas by shearing along the cut tip, the cracks propagate forward and facilitate releasing. Additionally, they implemented this pattern in a human-in-the-loop wearable device with high adhesion at the edges and easy removal after initial peeling. This device transmits the user's wrist motion to a mirrored robotic arm for manipulating various objects wirelessly (Figure 3Q).

## OPPORTUNITIES AND CHALLENGES

**Applications**—The large surface areas of robots, which are often underutilized, can be uniformly covered with kirigami metamaterials with adjustable properties and dynamic behaviors, such as variable stiffness,[65] distributed energy storage[44], and energy harvesting units[66]. The conformability of kirigami with embedded electronics facilitates the coverage of the curved skins of robots, turning them into adaptive displays.[67] Such electronic skins (E-skins), combined with the reconfigurable texture of kirigami, provide a means of social expression for human-robot interactions through visual signaling or controllable tactile experiences. Kinetic kirigami textures can also help robots blend into the background for camouflage[68]. The cuts in kirigami structures inherently render them permeable to light, heat, and fluids, allowing them to act as filters to regulate various processes such as body temperature[59] and sweat transport[60]. A kirigami skin can also be a mechanical end-effector, assisting the host robot in various tasks. For example, kirigami can facilitate locomotion and grasping by turning friction[30] and film adhesion[63] from a passive state to an active, spatially varying, controllable state. Opening of cuts in kirigami structures deviates mechanical stresses and maintains large low-strain areas in the structure for mounting delicate electronics and interconnections. The periodic arrangements of kirigami metamaterials enable easy and scalable assembly of modular units in different configurations to perform complex tasks. For instance, stretchable kirigami metamaterials become multistable in stretched state[65], allowing storage of bit information suitable for imbuing robots with mechanologic[69]. Integration of stimuli-responsive materials into kirigami structures that respond to heat, light, or chemical potentials can further enhance the autonomy of kirigami robots, enabling electronic-free control and environmental awareness[70]. Moreover, kirigami skins can provide a lightweight yet durable outer layer for robots, offering protection while maintaining maneuverability.

**Performance**—To evaluate the performance of kirigami-enabled robotic components, we need to define key metrics that relate their functions to the inherent characteristics of kirigami structures. For example, variations in the sensitivity of sensors, capacity of batteries, and the stroke of actuators with the deformability of kirigami sheets can be indicators for choosing them for specific applications. Mechanical properties such as stiffness, strength, strain at break, and fatigue life are other important parameters that define their longevity and safe working range. Quantifying texture reconfiguration of kirigami metasurfaces, for instance, the rotation angle of popups, can also affect robotic functions such as friction in locomotion, adhesion in grasping and wearables, convection rate in hydration and thermal regulation, and light absorption in solar cells.

**Limitations**—Another obstacle to incorporating kirigami into robotics lies in the inverse design of kirigami structures, which involves the transformation of 2D cut patterns into complex, animatable 3D structures[4,71]. These cuts must be meticulously crafted to preserve the intended robotic functionalities while avoiding material failure due to wear and tear or disruptions to embodied responsiveness. Moreover, the heterogeneity of cut morphology and the thinness of kirigami films pose additional technical burden for their integration into robots. The mechanical response of kirigami structures is highly nonlinear, and its numerical simulation, especially when active materials and contacts are involved, often requires in-house developments beyond the capabilities of existing finite element codes. Also, material and geometric nonlinearities make topology optimization of kirigami robots exceptionally challenging. Therefore, exploiting fast learning methods, such as self-supervised learning schemes[72], may accelerate the design of kirigami robots. Beyond design considerations, creating a streamlined fabrication process, such as layer-by-layer manufacturing[29], is imperative to expedite the production of multifunctional kirigami robots. Kirigami-enabled energy harvesters, batteries, and circuitry should be improved to balance energy consumption and ensure the continuous operation of kirigami robots. High-throughput and cost-effective production of kirigami robots with such embedded functionalities requires customized fabrication tools for multiplexing cutting, printing, and laminating techniques.[28,73]

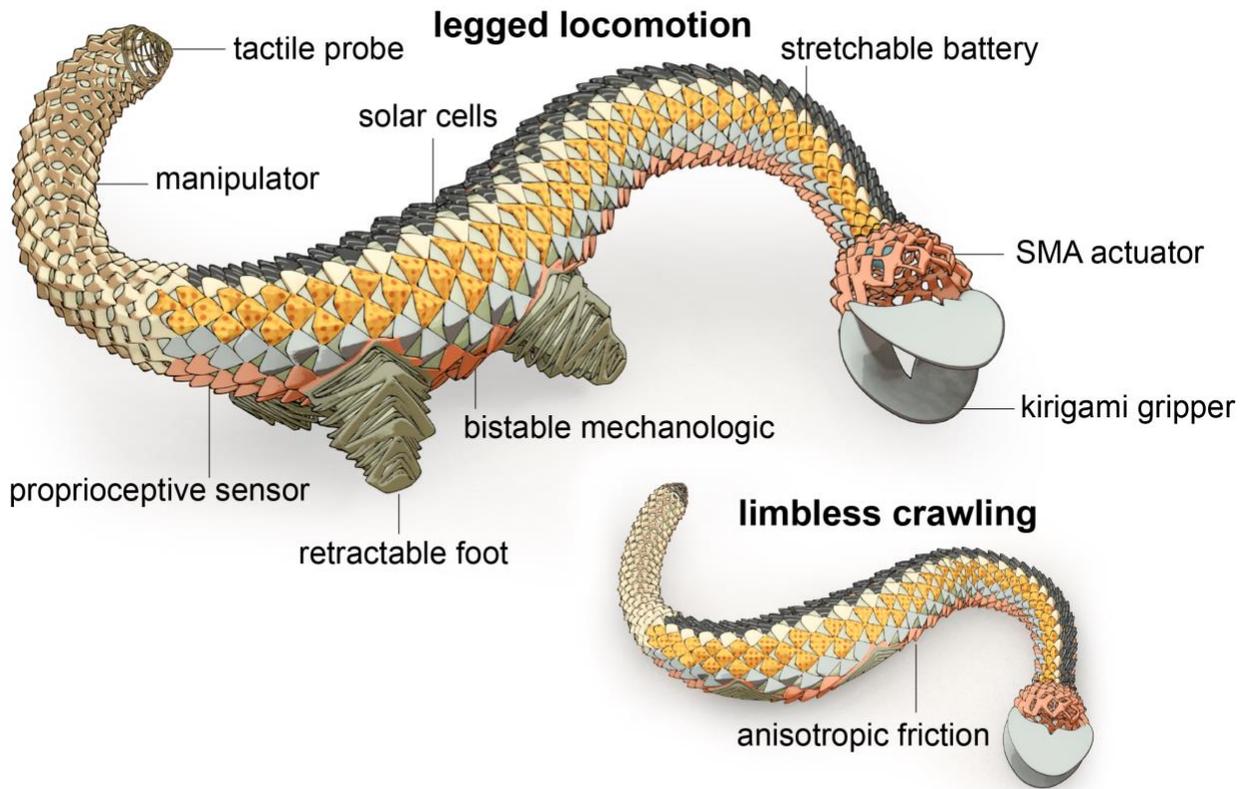

Figure 4. System integration and embodied intelligence in a conceptual kirigami robot
The robot is capable of locomotion and grasping that can switch from legged locomotion to limbless crawling with various integrated kirigami-enabled features. Illustrated by Sci-Fig Studio (www.sci-fig.com).

**Outlook**—To help envision the integration of different functionalities discussed in this perspective, we present a hypothetical robot shown in Figure 4, where kirigami-based robotic components are integrated at the system level. In this example, an SMA kirigami actuator enables programmed motions and grasping, kirigami resistive sensors provide proprioceptive feedback, and a metamaterial tactile sensor offers exteroceptive input for environmental awareness. Shape-morphing Li-ion kirigami batteries and stretchable kirigami perovskite solar cells allow power management. Based on these components, the robot can switch between legged locomotion and limbless crawling to tackle obstacles using the activation of a bistable mechanologic kirigami. Retractable feet enhance locomotion by overcoming barriers and when they are fully contracted, stretchable kirigami scales on the ventral side provide anisotropic friction for limbless crawling. The diverse integration strategies using kirigami can help enable new functionalities and improve existing abilities in robotics. Kirigami can also help simplify system designs, as one kirigami structure can fulfil multiple functions, such as actuation, perception, and structural support. We expect future endeavors in developing kirigami robots to converge toward system integration with augmented embodied intelligence, enhanced adaptability, and material sustainability.

## ACKNOWLEDGMENTS

This work was supported by the Villum Foundation through the Villum Young Investigator grant 37499. The image of the inflatable kirigami actuator shown in the *Graphical Abstract* is taken by Lishuai Jin.

## DECLARATION OF INTERESTS

The authors declare no competing interests.

## REFERENCES

1. Badalucco, L. (2000). Kirigami: The Art of Three-Dimensional Paper Cutting. New York: Sterling Publishing Company.
2. Rafsanjani, A., & Bertoldi, K. (2017). Buckling-induced kirigami. *Physical Review Letters*, 118(8), 084301. https://doi.org/10.1103/PhysRevLett.118.084301
3. Zhai, Z., Wu, L., & Jiang, H. (2021). Mechanical metamaterials based on origami and kirigami. *Applied Physics Reviews*, 8(4). https://doi.org/10.1063/5.0051088
4. Choi, G. P., Dudte, L. H., & Mahadevan, L. (2019). Programming shape using kirigami tessellations. *Nature Materials*, 18(9), 999-1004. https://doi.org/10.1038/s41563-019-0452-y
5. Jin, L., Forte, A. E., Deng, B., Rafsanjani, A., & Bertoldi, K. (2020). Kirigami-inspired inflatables with programmable shapes. *Advanced Materials*, 32(33), 2001863. https://doi.org/10.1002/adma.202001863
6. Babaee, S., Pajovic, S., Rafsanjani, A., Shi, Y., Bertoldi, K., & Traverso, G. (2020). Bioinspired kirigami metasurfaces as assistive shoe grips. *Nature Biomedical Engineering*, 4(8), 778-786. https://doi.org/10.1038/s41551-020-0564-3
7. Zhao, R., Lin, S., Yuk, H., & Zhao, X. (2018). Kirigami enhances film adhesion. *Soft Matter*, 14(13), 2515-2525. https://doi.org/10.1039/C7SM02338C
8. Hwang, D. G., Trent, K., & Bartlett, M. D. (2018). Kirigami-inspired structures for smart adhesion. *ACS applied materials & interfaces*, 10(7), 6747-6754. https://doi.org/10.1021/acsami.7b18594
9. Ai, C., Chen, Y., Xu, L., Li, H., Liu, C., Shang, F., Xia, Q. & Zhang, S. (2021). Current Development on Origami/Kirigami-Inspired Structure of Creased Patterns toward Robotics. *Advanced Engineering Materials*, 23(10), 2100473. https://doi.org/10.1002/adem.202100473

10. Brooks, A. K., Chakravarty, S., Ali, M., & Yadavalli, V. K. (2022). Kirigami-inspired biodesign for applications in healthcare. *Advanced Materials*, 34(18), 2109550. https://doi.org/10.1002/adma.202109550
11. Tao, J., Khosravi, H., Deshpande, V., & Li, S. (2023). Engineering by Cuts: How Kirigami Principle Enables Unique Mechanical Properties and Functionalities. *Advanced Science*, 10(1), 2204733. https://doi.org/10.1002/advs.202204733
12. Jin, L., & Yang, S. (2023). Engineering Kirigami Frameworks Toward Real-World Applications. *Advanced Materials*, 2308560. https://doi.org/10.1002/adma.202308560
13. Sareh, S., & Rossiter, J. (2012). Kirigami artificial muscles with complex biologically inspired morphologies. *Smart Materials and Structures*, 22(1), 014004. https://doi.org/10.1088/0964-1726/22/1/014004
14. Rafsanjani, A., Bertoldi, K., & Studart, A. R. (2019). Programming soft robots with flexible mechanical metamaterials. *Science Robotics*, 4(29), eaav7874. https://doi.org/10.1126/scirobotics.aav7874
15. Hong, Y., Zhao, Y., Berman, J., Chi, Y., Li, Y., Huang, H., & Yin, J. (2023). Angle-programmed tendril-like trajectories enable a multifunctional gripper with ultradelicacy, ultrastrength, and ultraprecision. *Nature Communications*, 14(1), 4625. https://doi.org/10.1038/s41467-023-39741-6
16. Fu, H., Nan, K., Bai, W., Huang, W., Bai, K., Lu, L., Zhou, C., Liu, Y., Liu, F., Wang, J. and Han, M. (2018). Morphable 3D mesostructures and microelectronic devices by multistable buckling mechanics. *Nature Materials*, 17(3), pp.268-276. https://doi.org/10.1038/s41563-017-0011-3
17. Chen, S., Liu, Z., Du, H., Tang, C., Ji, C.Y., Quan, B., Pan, R., Yang, L., Li, X., Gu, C. and Zhang, X. (2021). Electromechanically reconfigurable optical nano-kirigami. *Nature communications*, 12(1), p.1299. https://doi.org/10.1038/s41467-021-21565-x
18. Dias, M. A., McCarron, M. P., Rayneau-Kirkhope, D., Hanakata, P. Z., Campbell, D. K., Park, H. S., & Holmes, D. P. (2017). Kirigami actuators. *Soft Matter*, 13(48), 9087-9092. https://doi.org/10.1039/C7SM01693J
19. Shyu, T. C., Damasceno, P. F., Dodd, P. M., Lamoureux, A., Xu, L., Shlian, M., … Kotov, N. A. (2015). A kirigami approach to engineering elasticity in nanocomposites through patterned defects. *Nature Materials*, 14(8), 785-789. https://doi.org/10.1038/nmat4327
20. Blees, M. K., Barnard, A. W., Rose, P. A., Roberts, S. P., McGill, K. L., Huang, P. Y., … McEuen, P. L. (2015). Graphene kirigami. *Nature*, 524(7564), 204-207. https://doi.org/10.1038/nature14588
21. An, N., Domel, A. G., Zhou, J., Rafsanjani, A., & Bertoldi, K. (2020). Programmable hierarchical kirigami. *Advanced Functional Materials*, 30(6), 1906711. https://doi.org/10.1002/adfm.201906711
22. Seyidoğlu, B., Babu, S. P. M., & Rafsanjani, A. (2023, April). Reconfigurable kirigami skins steer a soft robot. In *2023 IEEE International Conference on Soft Robotics (RoboSoft)* (pp. 1-6). IEEE. https://doi.org/10.1109/RoboSoft55895.2023.10121995
23. Rafsanjani, A., Jin, L., Deng, B., & Bertoldi, K. (2019). Propagation of pop ups in kirigami shells. *Proceedings of the National Academy of Sciences*, 116(17), 8200-8205. https://doi.org/10.1073/pnas.1817763116

24. Tang, Y., Lin, G., Yang, S., Yi, Y. K., Kamien, R. D., & Yin, J. (2017). Programmable kiri-kirigami metamaterials. *Advanced Materials*, 29(10), 1604262. https://doi.org/10.1002/adma.201604262
25. Isobe, M., & Okumura, K. (2016). Initial rigid response and softening transition of highly stretchable kirigami sheet materials. *Scientific Reports*, 6(1), 24758. https://doi.org/10.1038/srep24758
26. Ma, Y., Feng, X., Rogers, J. A., Huang, Y., & Zhang, Y. (2017). Design and application of 'J-shaped' stress–strain behavior in stretchable electronics: a review. *Lab on a Chip*, 17(10), 1689-1704. https://doi.org/10.1039/C7LC00289K
27. Peter, C. (2011). Robotics, vision and control: fundamental algorithms in MATLAB.
28. Shigemune, H., Maeda, S., Hara, Y., Koike, U., & Hashimoto, S. (2015, September). Kirigami robot: Making paper robot using desktop cutting plotter and inkjet printer. In *2015 IEEE/RSJ International Conference on Intelligent Robots and Systems (IROS)* (pp. 1091-1096). IEEE. https://doi.org/10.1109/IROS.2015.7353506
29. Firouzeh, A., Higashisaka, T., Nagato, K., Cho, K., & Paik, J. (2020). Stretchable kirigami components for composite meso-scale robots. *IEEE Robotics and Automation Letters*, 5(2), 1883-1890. https://doi.org/10.1109/LRA.2020.2969924
30. Rafsanjani, A., Zhang, Y., Liu, B., Rubinstein, S. M., & Bertoldi, K. (2018). Kirigami skins make a simple soft actuator crawl. *Science Robotics*, 3(15), eaar7555. https://doi.org/10.1126/scirobotics.aar7555
31. Belding, L., Baytekin, B., Baytekin, H. T., Rothemund, P., Verma, M. S., Nemiroski, A., ... Whitesides, G. M. (2018). Slit tubes for semisoft pneumatic actuators. *Advanced Materials*, 30(9), 1704446. https://doi.org/10.1002/adma.201704446
32. Chung, S., Coutinho, A., & Rodrigue, H. (2022). Manufacturing and Design of Inflatable Kirigami Actuators. *IEEE Robotics and Automation Letters*, 8(1), 25-32. https://doi.org/10.1109/LRA.2022.3221318
33. Hwang, D. G., & Bartlett, M. D. (2018). Tunable mechanical metamaterials through hybrid kirigami structures. *Scientific Reports*, 8(1), 3378. https://doi.org/10.1038/s41598-018-21479-7
34. Duhr, P., Meier, Y. A., Damanpack, A., Carpenter, J., Studart, A. R., Rafsanjani, A., & Demirörs, A. F. (2023). Kirigami Makes a Soft Magnetic Sheet Crawl. *Advanced Science*, 10(25), 2301895. https://doi.org/10.1002/advs.202301895
35. Chen, J., Jiang, J., Weber, J., Gimenez-Pinto, V., & Peng, C. (2023). Shape Morphing by Topological Patterns and Profiles in Laser-Cut Liquid Crystal Elastomer Kirigami. *ACS Applied Materials & Interfaces*, 15(3), 4538-4548. https://doi.org/10.1021/acsami.2c20295
36. Tabassian, R., Mahato, M., Nam, S., Nguyen, V. H., Rajabi-Abhari, A., & Oh, I. K. (2021). Electro-Active and Photo-Active Vanadium Oxide Nanowire Thermo-Hygroscopic Actuators for Kirigami Pop-up. *Advanced Science*, 8(23), 2102064. https://doi.org/10.1002/advs.202102064
37. Zhu, H., Wang, Y., Ge, Y., Zhao, Y., & Jiang, C. (2022). Kirigami-Inspired Programmable Soft Magnetoresponsive Actuators with Versatile Morphing Modes. *Advanced Science*, 9(32), 2203711. https://doi.org/10.1002/advs.202203711
38. Cheng, Y. C., Lu, H. C., Lee, X., Zeng, H., & Priimagi, A. (2020). Kirigami-based light-induced shape-morphing and locomotion. *Advanced Materials*, 32(7), 1906233. https://doi.org/10.1002/adma.201906233

39. Rich, S. I., Wood, R. J., & Majidi, C. (2018). Untethered soft robotics. *Nature Electronics*, 1(2), 102-112. https://doi.org/10.1038/s41928-018-0024-1
40. Zheng, C., Oh, H., Devendorf, L., & Do, E. Y. L. (2019, June). Sensing kirigami. In *Proceedings of the 2019 on Designing interactive Systems Conference* (pp. 921-934). https://doi.org/10.1145/3322276.3323689
41. Chen, J., Shi, Y., Ying, B., Hu, Y., Gao, Y., Luo, S., & Liu, X. (2024). Kirigami-enabled stretchable laser-induced graphene heaters for wearable thermotherapy. *Materials Horizons*. https://doi.org/10.1039/D3MH01884A
42. Truby, R. L., Della Santina, C., & Rus, D. (2020). Distributed proprioception of 3D configuration in soft, sensorized robots via deep learning. *IEEE Robotics and Automation Letters*, 5(2), 3299-3306. https://doi.org/10.1109/LRA.2020.2976320
43. Wang, X. Q., Chan, K. H., Cheng, Y., Ding, T., Li, T., Achavananthadith, S., ... Ho, G. W. (2020). Somatosensory, light-driven, thin-film robots capable of integrated perception and motility. *Advanced Materials*, 32(21), 2000351. https://doi.org/10.1002/adma.202000351
44. Aubin, C. A., Gorissen, B., Milana, E., Buskohl, P. R., Lazarus, N., Slipher, G. A., ... Shepherd, R. F. (2022). Towards enduring autonomous robots via embodied energy. *Nature*, 602(7897), 393-402. https://doi.org/10.1038/s41586-021-04138-2
45. Kim, M. H., Nam, S., Oh, M., Lee, H. J., Jang, B., & Hyun, S. (2022). Bioinspired, shape-morphing scale battery for untethered soft robots. *Soft Robotics*, 9(3), 486-496. https://doi.org/10.1089/soro.2020.0175
46. Luo, L., Han, J., Xiong, Y., Huo, Z., Dan, X., Yu, J., Yang, J., Li, L., Sun, J., Xie, X., Wang, Z.L., Sun, Q. (2022). Kirigami interactive triboelectric mechanologic. *Nano Energy*, 99, 107345. https://doi.org/10.1016/j.nanoen.2022.107345
47. Hwang, D., Barron III, E. J., Haque, A. T., & Bartlett, M. D. (2022). Shape morphing mechanical metamaterials through reversible plasticity. *Science Robotics*, 7(63), eabg2171. https://doi.org/10.1126/scirobotics.abg2171
48. Ham, J., Han, A. K., Cutkosky, M. R., & Bao, Z. (2022). UV-laser-machined stretchable multi-modal sensor network for soft robot interaction. *npj Flexible Electronics*, 6(1), 94. https://doi.org/10.1038/s41528-022-00225-0
49. Janbaz, S., & Coulais, C. (2024). Diffusive kinks turn kirigami into machines. *Nature Communications*, 15(1), 1255. https://doi.org/10.1038/s41467-024-45602-7
50. Yang, Y., Vella, K., & Holmes, D. P. (2021). Grasping with kirigami shells. *Science Robotics*, 6(54), eabd6426. https://doi.org/10.1126/scirobotics.abd6426
51. Buzzatto, J., Mariyama, T., MacDonald, B. A., & Liarokapis, M. (2023). A Soft, Multi-Layer, Kirigami Inspired Robotic Gripper with a Compact, Compression-Based Actuation System. *IEEE/RSJ International Conference on Intelligent Robots and Systems (IROS)* (pp. 4488-4495). https://doi.org/10.1109/IROS55552.2023.10341893
52. Chaudhary, G., Niu, L., Han, Q., Lewicka, M., & Mahadevan, L. (2023). Geometric mechanics of ordered and disordered kirigami. *Proceedings of the Royal Society A*, 479(2274), 20220822. https://doi.org/10.1098/rspa.2022.0822
53. Hong, Y., Chi, Y., Wu, S., Li, Y., Zhu, Y., & Yin, J. (2022). Boundary curvature guided programmable shape-morphing kirigami sheets. *Nature Communications*, 13(1), 530. https://doi.org/10.1038/s41467-022-28187-x

54. Branyan, C., Hatton, R. L., & Mengüç, Y. (2020). Snake-inspired kirigami skin for lateral undulation of a soft snake robot. *IEEE Robotics and Automation Letters*, 5(2), 1728-1733. https://doi.org/10.1109/LRA.2020.2969949
55. Branyan, C., Rafsanjani, A., Bertoldi, K., Hatton, R. L., & Mengüç, Y. (2022). Curvilinear kirigami skins let soft bending actuators slither faster. *Frontiers in Robotics and AI*, 9, 872007. https://doi.org/10.3389/frobt.2022.872007
56. Liu, B., Ozkan-Aydin, Y., Goldman, D. I., & Hammond, F. L. (2019, April). Kirigami skin improves soft earthworm robot anchoring and locomotion under cohesive soil. In *2019 2nd IEEE International Conference on Soft Robotics (RoboSoft)* (pp. 828-833). IEEE. https://doi.org/10.1109/ROBOSOFT.2019.8722821
57. Tang, Y., Li, Y., Hong, Y., Yang, S., & Yin, J. (2019). Programmable active kirigami metasheets with more freedom of actuation. *Proceedings of the National Academy of Sciences*, 116(52), 26407-26413. https://doi.org/10.1073/pnas.1906435116
58. Li, Y., Halwah, A., Bhuiyan, S. R., & Yao, S. (2023). Bio-Inspired Untethered Robot-Sensor Platform for Minimally Invasive Biomedical Sensing. *ACS Applied Materials & Interfaces*, 15(50), 58839-58849. https://doi.org/10.1021/acsami.3c13425
59. Chen, T. H., Hong, Y., Fu, C. T., Nandi, A., Xie, W., Yin, J., & Hsu, P. C. (2023). A kirigami-enabled electrochromic wearable variable-emittance device for energy-efficient adaptive personal thermoregulation. *PNAS Nexus*, 2(6), pgad165. https://doi.org/10.1093/pnasnexus/pgad165
60. Wang, W., Yao, L., Cheng, C.Y., Zhang, T., Atsumi, H., Wang, L., Wang, G., Anilionyte, O., Steiner, H., Ou, J., Zhou, K. (2017). Harnessing the hygroscopic and biofluorescent behaviors of genetically tractable microbial cells to design biohybrid wearables. *Science Advances*, 3(5), e1601984. https://doi.org/10.1126/sciadv.1601984
61. Uetani, K., Kasuya, K., Wang, J., Huang, Y., Watanabe, R., Tsuneyasu, S., Satoh, T., Koga, H. and Nogi, M. (2021). Kirigami-processed cellulose nanofiber films for smart heat dissipation by convection. *npg Asia Materials*, 13(1), p.62. https://doi.org/10.1038/s41427-021-00329-5
62. Evke, E. E., Meli, D., & Shtein, M. (2019). Developable rotationally symmetric Kirigami-based structures as sensor platforms. *Advanced Materials Technologies*, *4*(12), 1900563. https://doi.org/10.1002/admt.201900563
63. Hwang, D., Lee, C., Yang, X., Pérez-González, J.M., Finnegan, J., Lee, B., Markvicka, E.J., Long, R. and Bartlett, M.D. (2023). Metamaterial adhesives for programmable adhesion through reverse crack propagation. *Nature Materials*, *22*(8), pp.1030-1038. https://doi.org/10.1038/s41563-023-01577-2
64. Hu, Y., & Hoffman, G. (2023). What Can a Robot's Skin Be? Designing Texture-changing Skin for Human–Robot Social Interaction. *ACM Transactions on Human-Robot Interaction*, 12(2), 1-19. https://doi.org/10.1145/3532772
65. Yang, Y., Dias, M. A., & Holmes, D. P. (2018). Multistable kirigami for tunable architected materials. *Physical Review Materials*, 2(11), 110601. https://doi.org/10.1103/PhysRevMaterials.2.110601
66. Lamoureux, A., Lee, K., Shlian, M., Forrest, S. R., & Shtein, M. (2015). Dynamic kirigami structures for integrated solar tracking. *Nature Communications*, 6(1), 8092. https://doi.org/10.1038/ncomms9092

67. Deng, Y., Xu, K., Jiao, R., Liu, W., Cheung, Y.K., Li, Y., Wang, X., Hou, Y., Hong, W. and Yu, H. (2024). Rotating square tessellations enabled stretchable and adaptive curved display. *npj Flexible Electronics*, *8*(1), p.4. https://doi.org/10.1038/s41528-023-00291-y
68. Lai, X., Peng, J., Cheng, Q., Tomsia, A.P., Zhao, G., Liu, L., Zou, G., Song, Y., Jiang, L. and Li, M. (2021). Bioinspired color switchable photonic crystal silicone elastomer kirigami. *Angewandte Chemie*, 133(26), 14428-14433. https://doi.org/10.1002/anie.202103045
69. Yang, Y., Feng, J., & Holmes, D. P. (2024). Mechanical Computing with Transmissive Snapping of Kirigami Shells. *Advanced Functional Materials,* 2403622. https://doi.org/10.1002/adfm.202403622
70. He, Q., Yin, R., Hua, Y., Jiao, W., Mo, C., Shu, H., & Raney, J. R. (2023). A modular strategy for distributed, embodied control of electronics-free soft robots. *Science Advances*, *9*(27), eade9247. https://doi.org/10.1126/sciadv.ade9247
71. Chen, T., Panetta, J., Schnaubelt, M., & Pauly, M. (2021). Bistable auxetic surface structures. *ACM Transactions on Graphics (TOG)*, 40(4), 1-9. https://doi.org/10.1145/3450626.3459940
72. Liu, R., Liang, J., Sudhakar, S., Ha, H., Chi, C., Song, S., & Vondrick, C. (2024). PaperBot: Learning to Design Real-World Tools Using Paper. *arXiv preprint* arXiv:2403.09566. https://arxiv.org/html/2403.09566v1
73. Wang, G., Cheng, T., Do, Y., Yang, H., Tao, Y., Gu, J., An, B. and Yao, L. (2018, April). Printed paper actuator: A low-cost reversible actuation and sensing method for shape changing interfaces. In *Proceedings of the 2018 CHI Conference on Human Factors in Computing Systems* (pp. 1-12). https://doi.org/10.1145/3173574.3174143